\documentclass{article}

    \usepackage[final]{neurips_2023}

\usepackage[utf8]{inputenc} %
\usepackage[T1]{fontenc}    %
\usepackage[colorlinks]{hyperref}       %
\usepackage{url}            %
\usepackage{booktabs}       %
\usepackage{amsfonts}       %
\usepackage{nicefrac}       %
\usepackage{microtype}      %
\usepackage{xcolor}         %
\usepackage{graphicx}
\usepackage{amsmath}
\usepackage{subcaption}
\usepackage{caption}
\usepackage{listings}
\usepackage{float}
\usepackage{placeins}
\usepackage{spverbatim}
\usepackage{fancyvrb}
\usepackage{fvextra}
\usepackage{tikz}
\usepackage{svg}
\usetikzlibrary{calc,decorations.markings}
\usepackage{wrapfig}

\renewcommand{\paragraph}[1]{\smallskip\textbf{#1}~~}

\usepackage{xspace}
\newcommand{\eg}{\textit{e.g.\@}\xspace}
\newcommand{\ie}{\textit{i.e.\@}\xspace}

\usepackage[capitalise,nameinlink]{cleveref}
\crefname{section}{Sec.}{Secs.}
\crefname{algorithm}{Alg.}{Algs.}
\crefname{appendix}{App.}{Apps.}
\crefname{definition}{Def.}{Defs.}
\crefname{table}{Table}{Tables}

\renewcommand{\mid}{\,|\,}
\newcommand{\words}[1]{\protect\tikz[baseline=-.5ex]\protect\node[fill=black!10,inner sep=1pt,rounded corners=1pt]{\texttt{#1}};}

\newlength{\figurewidth}
\newlength{\figureheight}

\tikzset{->-/.style={decoration={
  markings,
  mark=at position .5 with {\arrow{>}}},postaction={decorate}}}

\title{Improving Discrete Diffusion Models \\ via Structured Preferential Generation}

\author{%
  Severi Rissanen \\
  Department of Computer Science\\
  Aalto University\\
  \texttt{severi.rissanen@aalto.fi} \\
  \And
  Markus Heinonen \\
  Department of Computer Science\\
  Aalto University\\
  \And
  Arno Solin \\
  Department of Computer Science\\
  Aalto University\\
}

\begin{document}

\maketitle

\begin{abstract}
  In the domains of image and audio, diffusion models have shown impressive performance. However, their application to discrete data types, such as language, has often been suboptimal compared to autoregressive generative models. This paper tackles the challenge of improving discrete diffusion models by introducing a structured forward process that leverages the inherent information hierarchy in discrete categories, such as words in text. Our approach biases the generative process to produce certain categories before others, resulting in a notable improvement in log-likelihood scores on the text8 dataset. This work paves the way for more advances in discrete diffusion models with potentially significant enhancements in performance.
  
\end{abstract}

\section{Discrete Diffusion Models}
Diffusion models can be described as hierarchical latent variable models that are trained with a variational lower bound objective on the marginal log-likelihood. The variational lower bound is formed using a predefined variational inference distribution 
    $q(\mathbf{z}_{1:T} \mid \mathbf{z}_0) = \prod_{t=1}^T q(\mathbf{z}_t \mid \mathbf{z}_{t-1})$,
where the latent trajectory $\mathbf{z}_{1:T}$ generates $T$ successive variations of the original data $\mathbf{z}_0$.
The generative model is defined as a Markov chain going in the opposite direction:
    $p_\theta(\mathbf{z}_{0:T}) = p(\mathbf{z}_T) \prod_{t=1}^T p_\theta(\mathbf{z}_{t-1} \mid \mathbf{z}_t)$,
where $\theta$ are neural network parameters and $p(\mathbf{z}_T)$ is some prior distribution that is tractable to sample from. The model is trained by maximizing the evidence lower bound:
\begin{align}
    &\log p_\theta(\mathbf{z}_0) \geq \mathbb{E}_q \left[ \log \frac{p_\theta(\mathbf{z}_{0:T})}{q(\mathbf{z}_{1:T}\mid\mathbf{z}_0)} \right] \nonumber \\
    &= {-}\mathbb{E}_q \bigg[ \mathrm{KL}(q(\mathbf{z}_T\mid\mathbf{z}_0) \,\|\, p(\mathbf{z}_T)) {+} \sum_{t=2}^{T-1} \mathrm{KL}(q(\mathbf{z}_{t-1}\mid\mathbf{z}_t, \mathbf{z}_0) \,\|\, p_\theta(\mathbf{z}_{t-1}\mid\mathbf{z}_t)) - \log p_\theta(\mathbf{z}_0\mid\mathbf{z}_1) \bigg],
\end{align}
where the first term is zero if we set the inference distribution, or `forward process', endpoint to $q(\mathbf{z}_T\mid\mathbf{z}_0) \approx p(\mathbf{z}_T)$. Usually all of these distributions are factorized with respect to the data dimensions (\eg, the pixels in an image). In the discrete diffusion framework first proposed by \cite{sohl2015deep} and later extended by \cite{hoogeboom2021argmax} and \cite{austin2021structured}, all of these distributions are categorical distributions (\eg, categorical distributions over the 256 possibilities over a single byte representing a pixel, or all the tokens in a text data set). In practice, the transitions are parameterized as $p_\theta(\mathbf{z}_{t-1}\mid \mathbf{z}_{t}) \propto \sum_{\mathbf{x}_0} q(\mathbf{z}_{t-1},\mathbf{z}_t \mid \mathbf{z}_0) p_\theta(\mathbf{z}_0\mid \mathbf{z}_t)$. 

\section{Motivation}

For text data, \citet{austin2021structured} suggested a structured forward process that induced a high probability of moving between semantically similar states. This seems intuitively sensible, since diffusion models for image data also transition to nearby states more probably than to far-away states. Their results were worse than with forward processes that had no particular structure \citep{austin2021structured}. This raises the question: What is the right way to add inductive biases and data structure to discrete diffusion models without hurting performance? In this paper, we explore \emph{generating some categories of the data distribution before others}, with the idea that different categories, e.g., token types in text, encode different types of information. This is also analogous to the fact that Gaussian diffusion models for images generate low frequencies before the higher ones.

\section{Related Work}

In the context of language modelling, there has been some research on non-diffusion based models that generate tokens in structured orders. \citet{gu2019insertionbased} propose an approach that treats the orders of generated tokens as latent variables in a variational inference framework, where both a searched adaptive order and different predefined strategies are applied. One of the approaches, also explored in \cite{ford2018importance}, is to separate generation into two stages where the most common tokens are generated first and the less common after that, and vice versa. The difference to our diffusion-based framework is that they use causal masking in the self-attention and generate the positions of the added tokens as well as the tokens themselves one at a time, whereas we have the possibility of updating all of the tokens in the sequence in parallel, the number of generation steps is not bound to the dimension of the data and the neural network architecture is not restricted in any way.

\section{Method}

We begin with the `absorbing' version of the discrete diffusion model from \cite{austin2021structured}, but modify it to generate some tokens before others. The typical `masking' diffusion model introduces an additional `absorbing' state in the discrete state space, moving original tokens to this state in a specific order until all are absorbed by step $T$. The generation then reveals the masked tokens.

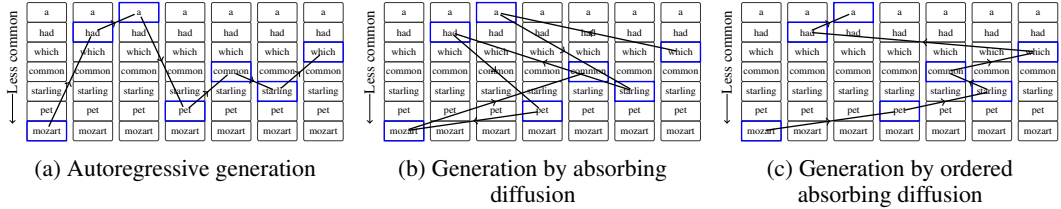
\begin{figure}[t!]
    \centering
    \captionsetup[subfigure]{justification=centering}
    \setlength{\figurewidth}{.06\textwidth}
    \setlength{\figureheight}{.03\textwidth}
    \newlength{\figsep}
    \setlength{\figsep}{.07\textwidth}
    
    \begin{subfigure}[t]{0.32\textwidth}
      \resizebox{\textwidth}{!}{%
        \begin{tikzpicture}[inner sep=0]

          \node[rotate=90,font=\small] (lab) at (.3\figurewidth,-3\figureheight) {Less common};
          \draw[->] (lab.west) -- ++(0,-1.5\figureheight);

          \foreach \j in {1,...,7} 
            \foreach \w [count=\i] in {a,had,which,common,starling,pet,mozart} 
              \node[text width=\figurewidth,minimum height=\figureheight,draw=black!80,align=center,font=\tiny,rounded corners=1pt] at (\j*\figsep,-\i*\figureheight) {\w};

          \foreach \i/\j [count=\n] in {1/7,2/2,3/1,4/6,5/4,6/5,7/3} 
              \node[text width=\figurewidth,minimum height=\figureheight,draw=blue,thick,align=center,font=\tiny] (node-\n) at (\i*\figsep,-\j*\figureheight) {};

          \foreach \from/\to [count=\n] in {1/2,2/3,3/4,4/5,5/6,6/7} 
              \draw[->-,thick,shorten >=.1cm,shorten <=.1cm] (node-\from.center) -> (node-\to.center);
                   
        \end{tikzpicture}}
        \caption{Autoregressive generation}
        \label{fig:autoreg_model}        
    \end{subfigure}
    \hfill
    \begin{subfigure}[t]{0.32\textwidth}
      \resizebox{\textwidth}{!}{%
        \begin{tikzpicture}[inner sep=0]

          \node[rotate=90,font=\small] (lab) at (.3\figurewidth,-3\figureheight) {Less common};
          \draw[->] (lab.west) -- ++(0,-1.5\figureheight);

          \foreach \j in {1,...,7} 
            \foreach \w [count=\i] in {a,had,which,common,starling,pet,mozart} 
              \node[text width=\figurewidth,minimum height=\figureheight,draw=black!80,align=center,font=\tiny,rounded corners=1pt] at (\j*\figsep,-\i*\figureheight) {\w};

          \foreach \i/\j [count=\n] in {7/3,3/1,6/5,2/2,4/6,1/7,5/4} 
              \node[text width=\figurewidth,minimum height=\figureheight,draw=blue,thick,align=center,font=\tiny] (node-\n) at (\i*\figsep,-\j*\figureheight) {};

          \foreach \from/\to [count=\n] in {1/2,2/3,3/4,4/5,5/6,6/7} 
              \draw[->-,thick,shorten >=.1cm,shorten <=.1cm] (node-\from.center) -> (node-\to.center);
                   
        \end{tikzpicture}}
        \caption{Generation by absorbing diffusion}
        \label{fig:autoreg_model}
    \end{subfigure}
    \hfill
    \begin{subfigure}[t]{0.32\textwidth}
        \resizebox{\textwidth}{!}{%
        \begin{tikzpicture}[inner sep=0]

          \node[rotate=90,font=\small] (lab) at (.3\figurewidth,-3\figureheight) {Less common};
          \draw[->] (lab.west) -- ++(0,-1.5\figureheight);

          \foreach \j in {1,...,7} 
            \foreach \w [count=\i] in {a,had,which,common,starling,pet,mozart} 
              \node[text width=\figurewidth,minimum height=\figureheight,draw=black!80,align=center,font=\tiny,rounded corners=1pt] at (\j*\figsep,-\i*\figureheight) {\w};

          \foreach \i/\j [count=\n] in {1/7,4/6,6/5,5/4,7/3,2/2,3/1} 
              \node[text width=\figurewidth,minimum height=\figureheight,draw=blue,thick,align=center,font=\tiny] (node-\n) at (\i*\figsep,-\j*\figureheight) {};

          \foreach \from/\to [count=\n] in {1/2,2/3,3/4,4/5,5/6,6/7} 
              \draw[->-,thick,shorten >=.1cm,shorten <=.1cm] (node-\from.center) -> (node-\to.center);
                   
        \end{tikzpicture}}
        \caption{Generation by ordered absorbing diffusion}
        \label{fig:autoreg_model}
    \end{subfigure}
    \caption{Three approaches to generating the expression: \words{mozart had a pet starling which}. The words are ordered top-down by how common they are, and the generation order is either left-to-right (a), random (b), or rare-words-first (c).}
\end{figure}

If we instead mask various categories at different steps in masking diffusion models, some categories are also generated earlier than others. This means we can introduce inductive biases, such as ordering tokens by their frequency in the original data. The forward distribution is defined with
\begin{align}
    q(\mathbf{z}_t=a \mid \mathbf{z}_0) = \delta_{a,M} m_t(\mathbf{z}_0) + \delta_{a,\mathbf{z}_0}(1-m_t(\mathbf{z}_0)).
\end{align}
Here, $M$ is the masking state and $m_t$ is the probability of having moved to the masking state by time $t$ from the initial category $z_0$. We recover the regular masking diffusion with the special case $m_t(\mathbf{z}_0) = m_t$, \ie, no dependence on the initial category itself. The $m_t$ functions leave us a lot of room for design. We propose a way to cut down this design space to a single schedule that allows for different tokens to be generated at different stages and where, on average, an equal amount of information is generated at each step. 

\paragraph{The mutual information schedule} \citet{austin2021structured} proposed to decrease the mutual information between the original data $x_0$ and $x_t$ at an equal amount at each step in the forward process as a heuristic. If $T$ is the total amount of forward steps, then at step $t$ the following equation should hold:
\begin{align}
	\frac{t}{T} = 1 - \frac{I(\mathbf{z}_0, \mathbf{z}_t)}{H(\mathbf{z}_0)} = \frac{H(\mathbf{z}_0, \mathbf{z}_t) - H(\mathbf{z}_t)}{H(\mathbf{z}_0)}.
\end{align}
For simplicity, the mutual informations and entropies are taken to be w.r.t.\ to the marginal distributions of all $\mathbf{z}_0, \mathbf{z}_t$ tokens. For the case of the standard mask diffusion and uniform diffusion, the schedule does not depend on the marginal distribution of $\mathbf{z}_0$, but for more complex cases such as ours, it does. The resulting formula for the forward process is:
\begin{align}
     \frac{t}{T} = \frac{\sum_{\mathbf{z}_0} p(\mathbf{z}_0) m_t(\mathbf{z}_0)\log\frac{p(\mathbf{z}_0)m_t(\mathbf{z}_0)}{\sum_{\mathbf{z}_0'}p(\mathbf{z}_0')m_t(\mathbf{z}_0')}}{\sum_{\mathbf{z}_0} p(\mathbf{z}_0)\log p(\mathbf{z}_0)}.
\end{align}

\paragraph{Narrowing down} The special case $m_t(\mathbf{z}_0) = m_t$ results in the regular mask diffusion with $\frac{t}{T} = m_t$. We choose $m_t(\mathbf{z}_0)$ to be such that in the limit of infinite steps $T\to \infty$, $m_t(\mathbf{z}_0)$ changes for a single $\mathbf{z}_0$ at a time. That is, for all but one $\mathbf{z}_0$, $m_t(\mathbf{z}_0)$ is either 0 or 1, and for a single $\mathbf{z}_0$, it is a monotonically increasing function between 0 and 1. Only one token is being destroyed at the same time. This, alongside the mutual information requirement, specifies the functions $m_t(\mathbf{z}_0)$ entirely. In practice, with a finite amount of timesteps $T$, we simply take snapshots of this idealized continuous process $\{m_t(\mathbf{z}_0)\}_{\mathbf{z}_0}$.

\paragraph{Choosing the order} 
We experiment with different token orderings. 1) Most frequent categories first 2) Least frequent categories first 3) A random order 4) An order based on information gain, as a proxy for finding conditional independence structures in data.
For the last one, we first estimate the marginal distribution of token types in sequences of fixed length from the data, and then calculate the information gain on this marginal distribution given that we observed some token type in the sequence. Formally:
\begin{align}
    \mathbb{E}_a \mathrm{IG}(X, a) = \mathbb{E}_a [H(X) - H(X \mid a)],
\end{align}
where $X$ is a random variable that gives the probability of observing different token types when sampling a single element from the sequence, and $a$ is the observation that a single example of a given token type exists in the sequence. The expectation $\mathbb{E}_a$ means that we consider both realizations of observing a token and not observing it in the sequence. The sequence length that we consider depends on the data set: For {\sc text8}, even though we generate sequences of length 200, we choose the information gain sequence length to be 10 since the amount of token types is low and it is likely that all token types are observed in longer sequences.

\section{Results}

We focus on text data for these experiments. Experiments done with simple Transformers without causal masking. 

\begin{wraptable}{r}{0.5\textwidth}
    \centering
    \begin{tabular}{c|c|c}
        & Valid ELBO & Train ELBO \\
        \hline
       Absorbing  &  0.181 & 0.4626\\
       Ordered  &  \textbf{0.129} & \textbf{0.3244}
    \end{tabular}
    \caption{Generating the tokens 'a' and 'b' before 'c', 'd', 'e' and 'f' improves results on a toy data set where the latter are conditionally independent of the former. }  
    \label{tab:toy_data}
\end{wraptable}

\paragraph{Toy data}
To illustrate why adding bias on the token order could improve the results, we experiment with the following toy data set: First, we sample every other token of a token sequence randomly with 50\% probability as \words{a}'s or \words{b}'s, and leave the other tokens blank, \eg, \words{a?b?b?a?b?a?$\dots$}. Second, the remaining blank tokens are filled with \words{c}:s,\words{d}:s,\words{e}:s and \words{f}:s deterministically depending on the combination of the surrounding two tokens with the rules \words{a?a}$\rightarrow$\words{aca}, \words{a?b}$\rightarrow$\words{acb}, \words{b?a}$\rightarrow$\words{bda} and \words{b?b}$\rightarrow$\words{beb}. Since there are correlations between the tokens, a standard absorbing-state model with a small enough amount of diffusion steps will produce mistakes due to the factorized nature of the reverse process $p_\theta(\mathbf{z}_{t-1} \mid \mathbf{z}_t)$. In contrast, a model that generates \words{a}:s and \words{b}:s first can, in principle, generalise perfectly in only two steps. We confirm this in a simple experiment where we train a mdoel to generate \words{a}:s and \words{b}:s before the others, and list the achieved losses in \cref{tab:toy_data}. 

The toy data set illustrates a mechanism through which a token-biased model can perform better: (Approximate) conditional independence of some token types given others. The question is then how to find such approximate conditional independences in data.

\begin{wrapfigure}{r}{0.5\textwidth}
    \centering
    \includegraphics[width=.5\textwidth]{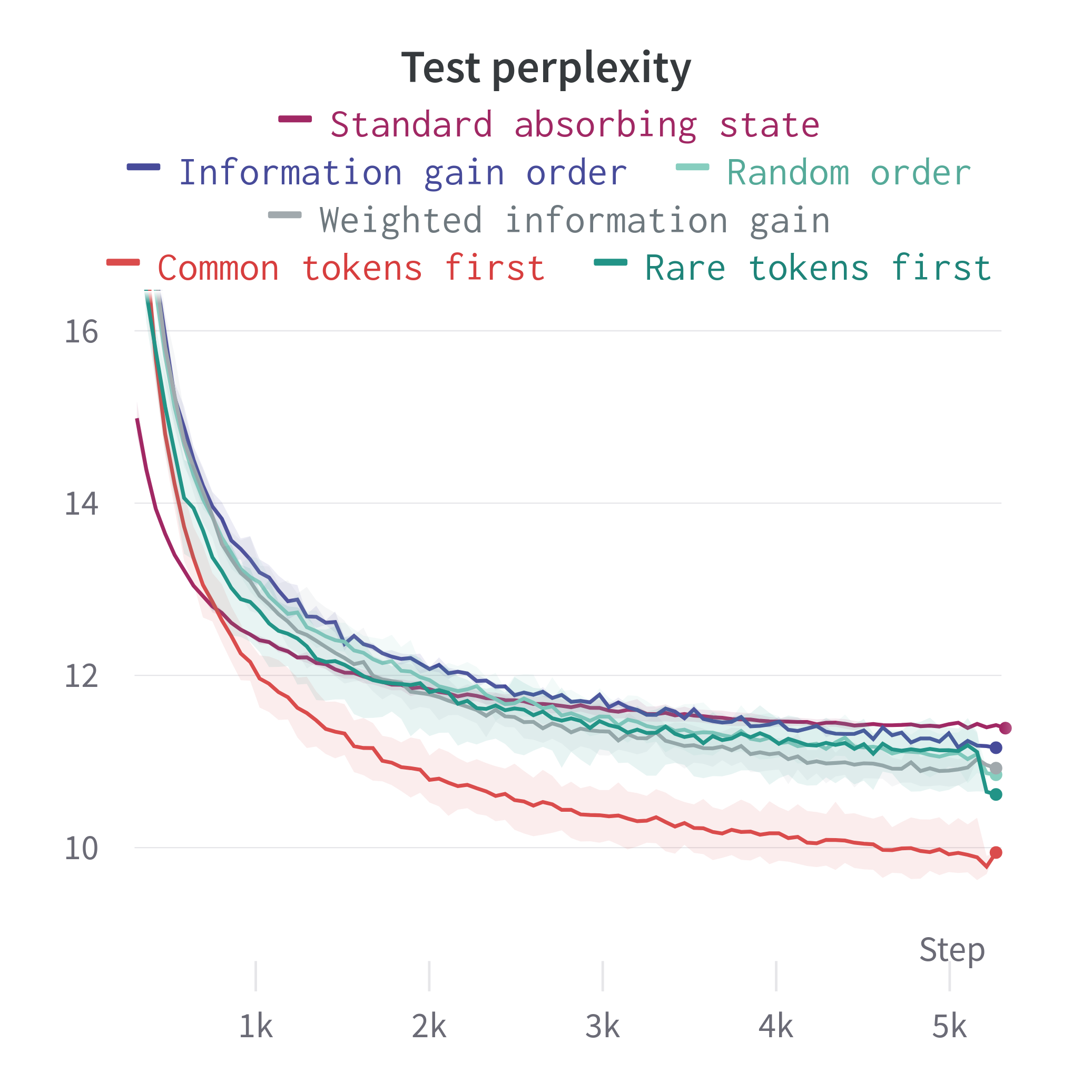}
    \caption{Test perplexities on {\sc text8} calculated with the ELBO as a proxy for the marginal likelihood.}\label{fig:text8}%
\end{wrapfigure}
\paragraph{\textsc{Text8}}
We next experiment on the {\sc text8} character-level data set. Here, we compare 5 different token orders, including on an information gain based ordering, but the MI schedule is skewed such that slightly more time is spent on the high information gain tokens. A visualization of the forward process with the common-first generation is shown in \cref{fig:text8_forward_visualization}. The models are 4-layer transformers trained on a single GPU with $T=1000$ diffusion steps.

\begin{wrapfigure}{R}{0.5\textwidth}
    \centering
    \includegraphics[width=0.5\textwidth]{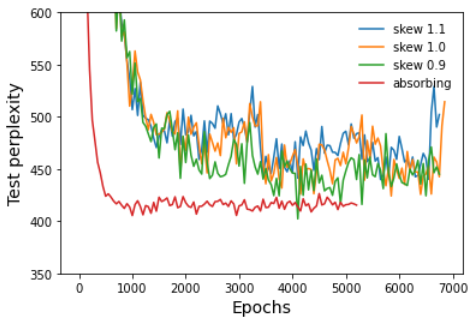}
    \caption{Test perplexities with different variants of the common-first model where tokens are grouped together with different strategies on {\sc Wikitext-2}, as well as the standard absorbing state model.}
    \label{fig:wikitext}
\end{wrapfigure}
\cref{fig:text8} shows the test perplexities with the different orderings across multiple runs for each ordering. The perplexities are estimated as $2^\mathrm{-ELBO}$, where the ELBO is normalized to per-character and expressed in log-base 2. The standard absorbing state model is also included for comparison. The common-first generation outperforms the other methods in text perplexity by a clear margin. Perhaps surprisingly, the information gain based ordering performs noticeably worse, although the skewed version seems to be slightly better on average. Generating the low-frequency tokens first performs about as badly as the random ordering. Another surprising observation is that all of the ordered methods perform slightly better than the standard absorbing state diffusion, on average, although the variance across runs is also clearly higher. Thus, especially the common-first method is able to fit better to the data, but training may be more challenging in practice.

\paragraph{\textsc{Wikitext}} %
In this example, our model is evaluated on the {\sc Wikitext-2} data set, which has a larger vocabulary (8300) compared to \textsc{text8} using a simple word-based tokenizer, leading to a challenge: most tokens will only have a chance of being generated in a single time step, leading to slow convergence in early experiments. Tokens were grouped into 20 blocks based on frequency on the training set, and the ordering was defined between those 20 blocks. We experimented with skewing the token frequencies for the blocking procedure with a parameter $\textrm{freq}^\alpha$, where $\alpha\in\{0.9,1,1.1\}$, to adjust the low-frequency blocks to be larger or smaller. We trained a transformer using common-first generation and a standard absorbing state model. As shown in \cref{fig:wikitext}, test perplexities were generally close but did not match the absorbing state model, suggesting a need for more careful design for larger data sets and complex token sets.

\section{Conclusion}
We have proposed and experimented on a new type of discrete diffusion model that generates some categories strictly before others, and discovered that it can clearly improve results with a simple text data set. Next steps include trying out more complex, non-character-level, text data sets as well as other types of data, such as graphs or segmentation maps. We believe that our model lays groundwork for a more systematic exploration of structured diffusion processes in discrete state spaces. %

\newpage

\bibliographystyle{plainnat}

\begin{thebibliography}{5}
\providecommand{\natexlab}[1]{#1}
\providecommand{\url}[1]{\texttt{#1}}
\expandafter\ifx\csname urlstyle\endcsname\relax
  \providecommand{\doi}[1]{doi: #1}\else
  \providecommand{\doi}{doi: \begingroup \urlstyle{rm}\Url}\fi

\bibitem[Austin et~al.(2021)Austin, Johnson, Ho, Tarlow, and Van
  Den~Berg]{austin2021structured}
Jacob Austin, Daniel~D Johnson, Jonathan Ho, Daniel Tarlow, and Rianne Van
  Den~Berg.
\newblock Structured denoising diffusion models in discrete state-spaces.
\newblock \emph{Advances in Neural Information Processing Systems},
  34:\penalty0 17981--17993, 2021.

\bibitem[Ford et~al.(2018)Ford, Duckworth, Norouzi, and
  Dahl]{ford2018importance}
Nicolas Ford, Daniel Duckworth, Mohammad Norouzi, and George Dahl.
\newblock The importance of generation order in language modeling.
\newblock In \emph{Proceedings of the 2018 Conference on Empirical Methods in
  Natural Language Processing}, pages 2942--2946, 2018.

\bibitem[Gu et~al.(2019)Gu, Liu, and Cho]{gu2019insertionbased}
Jiatao Gu, Qi~Liu, and Kyunghyun Cho.
\newblock {Insertion-based Decoding with Automatically Inferred Generation
  Order}.
\newblock \emph{Transactions of the Association for Computational Linguistics},
  7:\penalty0 661--676, 11 2019.

\bibitem[Hoogeboom et~al.(2021)Hoogeboom, Nielsen, Jaini, Forr{\'e}, and
  Welling]{hoogeboom2021argmax}
Emiel Hoogeboom, Didrik Nielsen, Priyank Jaini, Patrick Forr{\'e}, and Max
  Welling.
\newblock Argmax flows and multinomial diffusion: Learning categorical
  distributions.
\newblock \emph{Advances in Neural Information Processing Systems},
  34:\penalty0 12454--12465, 2021.

\bibitem[Sohl-Dickstein et~al.(2015)Sohl-Dickstein, Weiss, Maheswaranathan, and
  Ganguli]{sohl2015deep}
Jascha Sohl-Dickstein, Eric Weiss, Niru Maheswaranathan, and Surya Ganguli.
\newblock Deep unsupervised learning using nonequilibrium thermodynamics.
\newblock In \emph{International Conference on Machine Learning}, pages
  2256--2265. PMLR, 2015.

\end{thebibliography}

\newpage
\FloatBarrier

\lstset{%
  language=[LaTeX]TeX,
  backgroundcolor=\color{gray!15},
  basicstyle=\ttfamily,
  breaklines=true,
  columns=l,
  breakindent=0pt,
  breakatwhitespace=false
}

\begin{figure}
\begin{Verbatim}[breaklines=true, breakanywhere=true]
t=0 ployees it produced nothing but created jobs that would never have existed if one was only concerned with developing a real mine the world says the owner is an exploiter and the workers do all the real work he left the enterprise entirely in the hands of t
t=125 ploy??s?i??produc?d?no?hing?bu??cr?a??d?jobs??ha??would?n?v?r?hav???xis??d?if?on??was?only?conc?rn?d?wi?h?d?v?loping?a?r?al?min???h??world?says??h??own?r?is?an??xploi??r?and??h??work?rs?do?all??h??r?al?work?h??l?f???h???n??rpris???n?ir?ly?in??h??hands?of??
t=250 ploy??s?i??produc?d?no?hi?g?bu??cr????d?jobs??h???would???v?r?h?v???xis??d?if?o???w?s?o?ly?co?c?r??d?wi?h?d?v?lopi?g???r??l?mi????h??world?s?ys??h??ow??r?is?????xploi??r???d??h??work?rs?do??ll??h??r??l?work?h??l?f???h??????rpris?????ir?ly?i???h??h??ds?of??
t=375 pl?y??s????produc?d????h??g?bu??cr????d?j?bs??h???w?uld???v?r?h?v???x?s??d??f?o???w?s???ly?c??c?r??d?w??h?d?v?l?p??g???r??l?m?????h??w?rld?s?ys??h??ow??r??s?????xpl????r???d??h??w?rk?rs?d???ll??h??r??l?work?h??l?f???h??????rpr?s??????r?ly?????h??h??ds??f??
t=500 pl?y???????p??duc?d????h??g?bu??c?????d?j?b???h???w?uld???v???h?v???x????d??f?????w?s???ly?c??c????d?w??h?d?v?l?p??g??????l?m?????h??w??ld???ys??h???w???????????xpl????????d??h??w??k????d???ll??h?????l?w??k?h??l?f???h???????p???????????ly?????h??h??ds??f??
t=625 p??y???????p??duc?d???????g?bu??c?????d?j?b???????w?u?d???v?????v???x????d??f?????w??????y?c??c????d?w????d?v???p??g????????m????????w???d???y???????w???????????xp?????????d?????w??k????d???????????????w??k??????f???????????p????????????y???????????d???f??
t=750 p??y???????p??????????????g?b???????????j?b???????w???????v?????v???x???????f?????w??????y???????????w??????v???p??g????????m????????w???????y???????w???????????xp???????????????w??k????????????????????w??k??????f???????????p????????????y???????????????f??
t=875 ???y????????????????????????b???????????j?b???????w???????v?????v???x?????????????w??????y???????????w??????v????????????????????????w???????y???????w???????????x????????????????w??k????????????????????w??k???????????????????????????????y??????????????????
t=1000 ???????????????????????????????????????????????????????????????????????????????????????????????????????????????????????????????????????????????????????????????????????????????????????????????????????????????????????????????????????????????????????????????
\end{Verbatim}
\caption{Visualization of the forward process for text8 with the diffusion where common tokens are moved to the absorbing state first. The process starts out by diffusing 'e':s and spaces.}
\label{fig:text8_forward_visualization}
\end{figure}

\begin{figure}
\begin{Verbatim}[breaklines=true, breakanywhere=true]
t=1000 ????????????????????????????????????????????????????????????????????????????????????????????????????????????????????????????????????????????????????????????????????????????????????????????????????????????????????????????????????????????????????????????????
t=889 ? ???????? ???????????? ??? ??? ??????? ????? ???? ?? ??? ??????? ?? ???? ?? ??????? ?????? ?????? ??????? ???? ?????? ?????? ?? ???????? ???????????? ????? ?????????? ??? ???????? ?? ????????? ??? ??????? ??? ???? ???? ?? ???????? ?? ????? ??????????? ??
t=778 e e??????? ???? e??e??? ??? ??? ?e????? ???e? ??e? ?? ??e ??????? ?? ?e?? ?? ??????? ?????? ?? ??? e????e? ?e?e ?????? ?????? ?? ???????? ???????e? ?? ???e? ???????e?? ??? ???????? ?? ????????? ??? ?e????? ??? ?e?? ?e?? ?? ???? ??? ?? ????e ??????????? ??
t=667 e e???t??? t??? e??e??? ??? ?t? ?e?e??? ??te? ??e? ?? t?e ??????? ?? ?e?? ?? ??????? ?????? ?? t?e e????e? ?e?e ?e???? ?????? ?? ?e????t? ???????e? ?? ?t?e? ??????te?? t?e ???????? ?? ????????? ??? ?e??t?? ??? ?e?? ?e?? t? t??? ??? ?? ???te ???????t??? ??
t=556 e e???t??? t??? e??e??? a?? ?t? ?e?e?a? a?te? ??e? ?? t?e ?a????? a? ?e?? a? ?????a? ?????? a? t?e e????e? ?e?e ?e??an ???a?? an ?e????t? ????a??e? ?? ?t?e? ??a?a?te?? t?e ???????? ?? ???????n? a?? ?e??t?? t?? ?e?? ?e?? t? t??? ?a? ?n ???te t?a???at??? a?
t=444 e e???ti?n t?i? e??e??? an? it? ?e?e?a? a?te? ??e? ?n t?e ?a?i??? a? ?e?? a? ?????a? ?i?i?? a? t?e en???e? ?e?e ?e??an ???ain an ?e???it? ????a??e? in ?t?e? ??a?a?te?? t?e ?in??i?? ?? ??????in? an? ?en?t?? t?? ?e?? ?e?? t? t?i? ?a? in ??ite t?an??ati?n a?
t=333 e e???tion t?i? e??eror an? it? ?e?e?a? a?ter ??e? on t?e ?a?i??? a? ?e?? a? ?o???ar ?i?i?? a? t?e en?o?e? ?e?e ?er?an ???ain an ?e??rit? ????a??e? in ot?er ??a?a?te?? t?e ?in??i?? o? ?o??o?in? an? ?en?t?? t?o ?ero ?ero to t?i? ?a? in ??ite tran??ation a?
t=222 e e???tion this e??eror an? its se?era? a?ter ?se? on the ?a?i??? as ?e?? as ?o???ar si?i?? as the en?o?e? ?ere ?er?an h?rain an se??rit? s???a??e? in other ?hara?ters the ?in??is? o? ?o??o?in? an? ?en?ths t?o ?ero ?ero to this ?as in ??ite trans?ation as
t=111 e e???tion this e??eror and its se?eral a?ter used on the ?a?i?u? as ?ell as ?o?ular sicil? as the en?o?ed ?ere ?er?an hurain an securit? s?lla?led in other characters the lin?uis? o? ?ollo?in? and len?ths t?o ?ero ?ero to this ?as in ?uite translation as
t=0 e egyption this emperor and its several after used on the maximum as well as popular sicily as the enjoyed were german hurain an security syllabled in other characters the linguism of following and lengths two zero zero to this was in quite translation as
\end{Verbatim}
\caption{Visualization of the reverse process of the diffusion process where the common tokens are generated first. The trained model is a 12-layer transformer with about 10 million parameters.}
\label{fig:text8_forward_visualization}
\end{figure}

\begin{figure}
\begin{Verbatim}[breaklines=true, breakanywhere=true]
t=1000 ????????????????????????????????????????????????????????????????????????????????????????????????????????????????????????????????????????????????????????????????????????????????????????????????????????????????????????????????????????????????????????????????
t=875 ?????? ????? ????? ?? ??? ??????? ??????? ?????? ?????? ?? ? ????? ?????? ???? ??????? ????????????? ??? ????? ????????????? ???? ?????? ?? ??? ??? ????????? ???? ???? ????? ??? ???? ????? ????? ??? ??????? ???????? ??????????? ????????????? ??????? ?? ???
t=750 e??e? ???e? ????? ?? ??e ??????? ??? ??e ?e???? ?e???e ?? ? ????? ???e ? ???? ??? ??e ???e??e?e? ?? ??? ????? ???ee???????? ???? ????e? ?? ??e ??? ????????? ???? ??e? ????? ??? ???? ? e?? ???e? ??? ?e????e ???e??e? ?e?e? ????? ?ee??e????e?? ?e???e? ?e ??e
t=625 ea?e? ???e? ??a?? ?? t?e ????a?e ??? t?e ?e?t?? ?e??te ?? ? ????t ???e ? ???? ??? t?e ???e?te?e? ?? ??? ??t?? ???ee?at???e? ???t ????e? at t?e ??? ???????e? ???? ??e? ????? ?e? ???? ? e?? t??ee ??? ?e????e ??te??e? ?e?e? ????? ?ee??e????e?t ?e???e? ?e ??e
t=500 ea?e? ???e? ??a?? ?? t?e ?a??a?e ??n t?e ?eat?? ?e??te ?? a ????t ?a?e a ???? an? t?e ?n?e?te?e? ?? ??? ??t?? ?n?ee?at??ne? ???t ????e? at t?e ??? ??????ne? ?an? ??e? ???n? ne? ???? ? e?? t??ee ??? ?e????e ?nte??e? ?e?en ?a??? ?een?e????ent ?e???e? ?e ?ne
t=375 ea?e? ?i?e? ??a?? o? t?e ?a??a?e ?on t?e ?eat?? ?e?ote o? a ?i??t ?a?e a ??o? an? t?e in?e?te?e? ?? ?i? ??t?? in?ee?atione? ?o?t o???e? at t?e ?i? ?o???ine? ?an? o?e? ?o?n? ne? ??o? ? e?o t??ee ?o? ?e??i?e inte??e? ?e?en ?a?o? ?een?e??i?ent ?e?o?e? ?e one
t=250 ea?er ?i?es s?a?s o? the ?assa?e ?on t?e ?eat?? se?ote o? a ?irst ?a?e a ?ro? an? the in?e?te?e? ?? ?is ?ot?? in?ee?atione? ?ost o???e? at the ?is ?o???ine? ?an? o?er ?o?n? ne? ?ro? s ero three ?o? ser?i?e inte??es se?en ?a?or ?een?ersi?ent ?e?o?e? ?e one
t=125 ea?er ?i?es s?als o? the ?assa?e ?on the death? secote o? a ?irst ?a?e a ?ro? and the indecteded ?? his ?otld indee?ationed ?ost occ?ed at the his co??lined ?an? o?er ?ound ne? ?ro? s ero three co? ser?ice intelles se?en la?or ?eendersi?ent ?eco?ed ?e one
t=0 eaver gives smals of the kassage won the deathy secote of a first wave a from and the indecteded by his botld indeegationed most occued at the his complined many over found new from s ero three cox service intelles seven labor peendersiment becomed be one
\end{Verbatim}
\caption{Visualization of the reverse process of the diffusion process where the common tokens are generated first. The trained model is a 4-layer transformer with about 3 milllion parameters (also used for the experiments in Fig.\ref{fig:text8})}
\label{fig:text8_forward_visualization}
\end{figure}

\begin{figure}
\begin{Verbatim}[breaklines=true, breakanywhere=true]
t=1000 <mask> <mask> <mask> <mask> <mask> <mask> <mask> <mask> <mask> <mask> <mask> <mask> <mask> <mask> <mask> <mask> <mask> <mask> <mask> <mask>
t=889 <mask> , <mask> <mask> <mask> <mask> <mask> <mask> <mask> <mask> <mask> <mask> <mask> <mask> <mask> <mask> <mask> <mask> <mask> <mask>
t=778 <mask> , <mask> <mask> a <mask> to <mask> <mask> <mask> <mask> <mask> <mask> <mask> <mask> . <mask> <mask> <mask> <mask>
t=667 <mask> , <mask> <mask> a <mask> to <mask> <mask> <mask> <mask> <mask> as <mask> <mask> . <mask> <mask> <mask> '
t=556 <mask> , they <mask> a <mask> to their <mask> over new world as they <mask> . <mask> their <mask> '
t=444 <mask> , they <mask> a <mask> to their way over new world as they did . despite their men '
t=333 <mask> , they sent a <mask> to their way over new world as they did . despite their men '
t=222 arrived , they sent a <mask> to their way over new world as they did . despite their men '
t=111 arrived , they sent a <mask> to their way over new world as they did . despite their men '
t=0 arrived , they sent a platoon to their way over new world as they did . despite their men ' 
\end{Verbatim}
\caption{Visualization on the Wikitext data set of the reverse process of the diffusion process where the common tokens are generated first. The trained model is a 12-layer transformer with about 10 million parameters.}
\label{fig:text8_forward_visualization}
\end{figure}

\end{document}